\documentclass[10pt,twocolumn,letterpaper]{article}

\usepackage{iccv}
\usepackage{times}
\usepackage{epsfig}
\usepackage{graphicx}
\usepackage{amsmath}
\usepackage{amssymb}
\usepackage{authblk}

\usepackage[pagebackref=true,breaklinks=true,letterpaper=true,colorlinks,bookmarks=false]{hyperref}

\iccvfinalcopy 

\def\iccvPaperID{25} 

\ificcvfinal\fi

\begin{document}

\title{Explaining Vision and Language through Graphs of Events in Space and Time}


\author{Mihai Masala\textsuperscript{1,2} \hspace{1cm}
Nicolae Cudlenco\textsuperscript{1} \hspace{1cm}
Traian Rebedea\textsuperscript{2} \hspace{1cm}
Marius Leordeanu\textsuperscript{1,2}
\thanks{Primary contact: Marius Leordeanu at leordeanu@gmail.com}
\vspace{0.5mm}\\
\textsuperscript{\rm 1}
Institute of Mathematics of the Romanian Academy
\hspace{1.5cm}
\textsuperscript{\rm 2}University Politehnica of Bucharest 
}


\maketitle
\ificcvfinal\thispagestyle{empty}\fi

\begin{abstract}
Artificial Intelligence makes great advances today and starts to bridge the gap between vision and language. However, we are still far from understanding, explaining and controlling explicitly the visual content from a linguistic perspective, because we still lack a common explainable representation between the two domains. In this work we come to address this limitation and propose the Graph of Events in Space and Time (GEST), by which we can represent, create and explain, both visual and linguistic stories.
We provide a theoretical justification of our model and an experimental validation, which proves that GEST can bring a solid complementary value along powerful deep learning models. In particular, GEST can help improve at the content-level the generation of videos from text, by being easily incorporated into our novel video generation engine. Additionally, by using efficient graph matching techniques, the GEST graphs can also improve the comparisons between texts at the semantic level.
\end{abstract}

\section{Introduction} 

There is a considerable amount of research at the intersection of vision and language, 
such as image and video generation~\cite{reed2016generative,zhou2019text, li2018video, balaji2019conditional, wu2022nuwa, singer2022make, villegas2022phenaki}, captioning~\cite{gao2017video,zhou2018end,wang2018reconstruction} or visual question answering~\cite{antol2015vqa,lu2016hierarchical,zhong2020self}. However, we still lack an explainable model that can fully relate, constrain and control the connection between vision and language at the level of meaning and content. This limitation, which affects not only text-to-image/video models, but also Large Language Models~\cite{zhang2023language}, seriously impedes our way towards 
trustworthy and safe AI. We mention that, even in this work, we found state of the art text-to-video transformer models generating almost adult-only content for a simple, plain text such as: \textit{A woman goes to the bedroom}. 

\begin{figure}[h]
    \centering
    \includegraphics[width=0.5\textwidth]{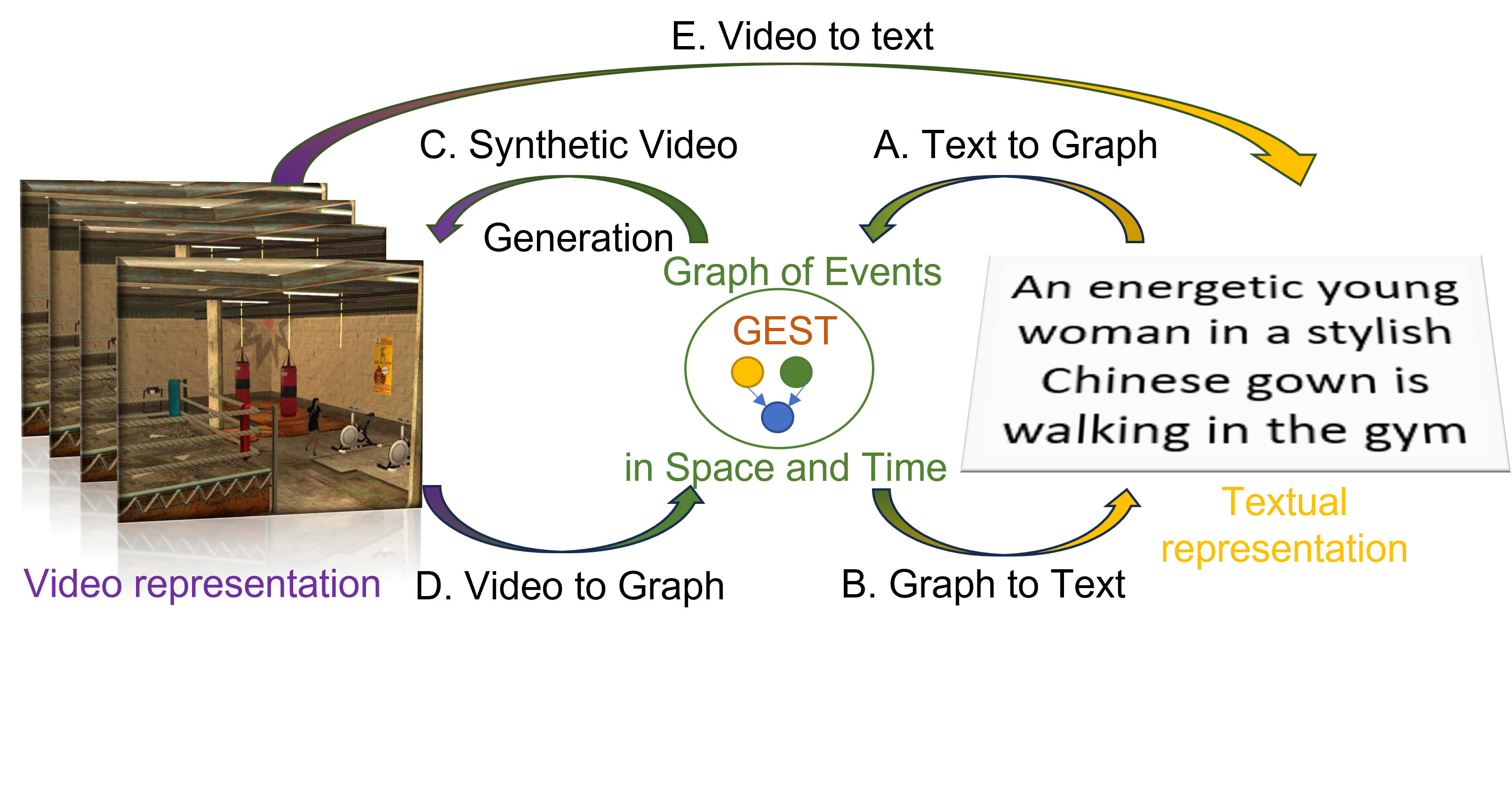}
    \caption{Functional overview of the proposed framework. GEST represents the central component, allowing for the preservation of the semantic content in an explainable form, as well as a seamless transition between different domains.}
    \label{fig:general}
\end{figure}

In this context, we introduce GEST, the Graph of Events in Space and Time, which provides an explicit spatio-temporal representation of stories as they appear in both videos and texts and can immediately relate, in an explainable way, the two domains. GEST provides a meaningful representation space, in which similarities between videos and texts can be computed at the level of semantic content.
GEST can also be used in the context of our specially designed video generation engine (Sec. \ref{sec:GEST_engine})
to produce videos that are rated higher in terms of content, both by human and automatic evaluations, than their video 
counterparts generated by state of the art text-to-video models (Sec. \ref{sec:vision_language_tests}).
Also, GEST graphs can be used for comparing the meaning of texts and improve over classic text similarity metrics or in combination with heavily trained state-of-the-art deep learning metrics
(Sec. \ref{sec:text_comparison}).
Graphs have been used to represent content in videos~\cite{sridhar2010relational,singh2017graph,brendel2011learning, yuan2017temporal,wang2018videos, cherian20222} or texts~\cite{lin1998dependency,zettlemoyer2012learning,mann1988rhetorical,christensen2013towards,banarescu2013abstract}, but not both as is the case for GEST.

\textbf{Main novel aspects of GEST} are: \textbf{1)} Nodes are events, which could represent (Sec. \ref{sec:GEST_model}) physical objects, simple actions or even complex activities and stories. \textbf{2)} Edges can represent any type of relation (temporal, spatial, semantic, as defined by any verb) between two events defined as nodes. \textbf{3)} Any GEST graph can always collapse into a node event, at a higher level of abstraction. Also, any event node can always be expanded into a GEST graph, from a lower level of abstraction. This is an essential property that allows GEST to have multiple layers of depth (see Fig. \ref{fig:gest_example_complex}). 

Another practical \textbf{contribution} of our work, is our novel video generation engine (Sec. \ref{sec:GEST_engine}), based on GEST, which can produce long and complex videos that preserve well semantic content, as validated by human and automatic evaluations. We will make the engine code and the videos generated for our experiments publicly available.

\section{GEST Model}
\label{sec:GEST_model}

The basic elements of GEST are the nodes, which represent events and the edges, which represent the way in which events interact. 

\textbf{GEST nodes:} represent events that could go from
simple actions (e.g. opening a door) to complex, high-level events (e.g. a political revolution), in terms of spatio-temporal extent, scale and semantics. They are usually confined to a specific time period (e.g. a precise millisecond or whole year) and space region (e.g. a certain room or entire country). Events could exist at different levels of semantics, ranging from simple physical contact (e.g. ``I touch the door handle'') to profoundly semantic ones (e.g. ``the government has fallen'' or ``John fell in love with physics''). 
Even physical objects are also events (e.g. John's car is represented by the event ``John's car exists''). Generally, any space-time entity could be a GEST event.

\textbf{GEST edges:} relate two events and can
define any kind of interaction between them, from simple temporal ordering (e.g. ``the door opened'' after ``I touched the door handle'') to highly semantic (e.g. ``the revolution'' caused ``the fall of the government'', or ``Einstein's discovery'' inspired ``John to fall in love with physics''). Generally, any verb that relates two events or entities could be a GEST edge.

\textbf{From a graph to a node and vice-versa:} A GEST graph essentially represents a story in space and time, which could be arbitrarily complex or simple. Even simple events can be explained by a GEST, since all events can be broken, at a sufficient level of detail, into simpler ones and their interactions (e.g. ``I open the door'' becomes a complex GEST if we describe in detail the movements of the hand and the mechanical components involved). At the same time, any GEST graph could be seen as a single event from a higher semantic and spatio-temporal scale (e.g. ``a political revolution'' could be both a GEST graph and a single event). Collapsing graphs into nodes ($Event \Leftarrow GEST$)
or expanding nodes into graphs ($GEST \Leftarrow Event$)
, gives GEST the possibility to have many levels of depth, as needed for complex visual and linguistic stories.

\begin{figure}
    \centering
    \includegraphics[width=\columnwidth]{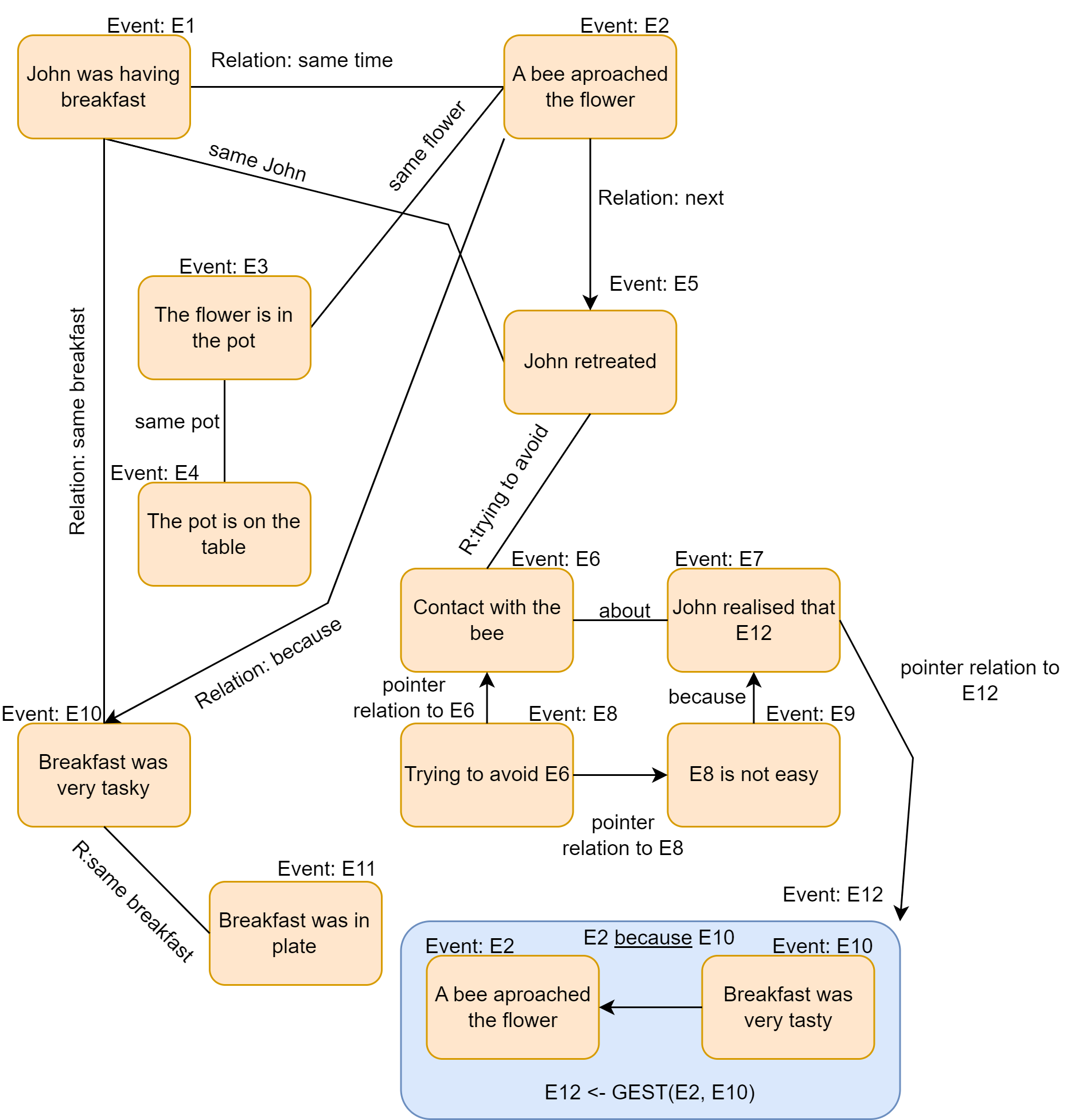}
    \caption{GEST graph explaining the following text: \textit{``John was having breakfast when a bee approached the flower in the pot on the table. Then he pulled back trying to avoid contact with the bee but he realized that it was not an easy attempt because she actually came because of the tasty food on his plate''.}}
    \label{fig:gest_example_complex}
\end{figure}

Going from a $GEST$ at a lower level to an event $E$ at a higher level ($E \Leftarrow GEST$) reminds of
how the attention mechanism is applied in Graph Neural Networks and Transformers \cite{vaswani2017attention}: the GEST graph acts as a function that aggregates information from nodes (events) $E_i$'s
at level $k$ and builds a higher level GEST representation, which 
further becomes an event at the next level $k+1$:

\[ E_i^{(k+1)} \Leftarrow GEST(E_1^k, E_2^k, ..., E_n^k)\]

In Fig. \ref{fig:gest_example_complex} we present our GEST representation, as it applies to a specific text.
In each event node, $E_i$, we encode an $action$, a list of $entities$ that are involved in the action, its $location$ and $timeframe$ and any additional $properties$. Note that an event can contain references (pointers) to other events, which define relations of type ``same X'' (e.g. ``same breakfast''). We also exemplify how the GEST of two connected events can collapse into a single event.

\begin{table}[h]
\begin{center}
\begin{tabular}{|l|c|c|c|c|}
\hline
\textbf{Method} & \textbf{Corr(\%)} &  \textbf{Acc(\%)} & \textbf{F} & \textbf{AUC(\%)} \\
\hline\hline

BLEU@4 & 24.45 & 75.52 & 0.28 & 52.65\\
METEOR & 58.48 & 84.23 & \underline{1.12} & 73.90\\
ROUGE & 51.11  & 83.40 & 0.72 & 68.92\\
SPICE & 59.42 & 84.65 & 1.04 & 74.43 \\
BERTScore & 57.39 & \underline{85.89} & 1.07 &\textbf{77.93}\\
\hline
GEST-SM& \textbf{61.70} & 84.65 & \textbf{1.20} & 75.47 \\
GEST-NGM& \underline{60.93} & \textbf{86.31} & 0.98 & \underline{76.75} \\

\hline
\end{tabular}
\end{center}
\caption{Comparing GEST 
representation power (coupled with graph matching similarity functions SM or NGM) and well-known text-to-text similarity methods (applied on texts from Videos-to-Paragraphs test set, on the task of separating texts describing the same video vs. texts from different videos). Corr - correlation, Acc - Accuracy, F - Fisher score and AUC - area under the precision-recall curve. Best values are in \textbf{bold}, second best \underline{underlined}.} 
\label{table:metrics_story_vs_classic}
\end{table}

\subsection{GEST for Textual Content Comparison}
\label{sec:text_comparison}
Next we verify experimentally that the GEST model can capture the semantics of language
by applying it to the task of text to text comparisons, in the context of video to text translation.
We use the Videos-to-Paragraphs dataset~\cite{bogolin2020hierarchical} that has 
multiple text descriptions for the same video. 
Starting from the given texts, we build ground truth GEST representations for the entire dataset as follows: we use a rule-based method to obtain initial GESTs from texts, represented in a specific string format that captures information in the nodes as well as their relationships. Next we check, correct and refine the automatically generated GESTs by human annotation. Note that we also tested with GhatGPT, which was able to produce mostly valid GESTs by learning from a few human examples. 

We seek to find how useful is GEST in deciding if two texts stem from the same video or not. 
Basically, instead of comparing texts, we move the comparison in the GEST space in which we define a similarity function using graph matching. In
We use as graph matching methods the classic 
Spectral Matching (SM)~\cite{leordeanu2005spectral} and the recent Neural Graph Matching (NGM)~\cite{wang2021neural}. 
For both algorithms, the affinity matrix is build using node and edge level similarity functions based on pre-trained GloVe~\cite{pennington2014glove} word embeddings. Two nodes are as similar as are their components (e.g. action, entities), while edge-level similarity uses the relation type
defined by the edge (e.g. causality, temporal ordering, etc.) along with the
similarity of the nodes they connect.

In Tab.~\ref{table:metrics_story_vs_classic} we present comparisons of GEST+graph matching similarity vs.
other well-known text similarity metrics, which demonstrate that GEST is capable to capture semantic content.
In Tab.~\ref{table:metrics_story_learned} we investigate whether graph matching in GEST space can be combined 
with state-of-the-art highly trained text similarity metrics such as BLEURT~\cite{sellam-etal-2020-bleurt}. We combine each pair of similarity metrics (BLEURT + X) in linear way, to ensure that if a performance gain exists, it is less likely to be due to the combination method and more due to the additional metric. In this setting GEST graphs are learned by finetuning a GPT-3 model (\textit{text-curie-001}), with raw text as input and ground truth GEST as output, on the Videos-to-Paragraphs train set. Note that the combination of BLEURT with graph matching in the GEST space consistently increases the performance over BLEURT (which is not always the case for other metrics) and by the largest margin.


\begin{table}[h]
\begin{center}    
\begin{tabular}{|l|c|c|c|c|}
\hline
\textbf{Method} & \textbf{Corr(\%)} &  \textbf{Acc(\%)} & \textbf{F} & \textbf{AUC(\%)} \\
\hline\hline

BLEURT & 70.93 & 90.04 & 2.03 & 88.02 \\
\hline
+BLEU@4 & 70.93 & 90.04 & 2.03 & 88.04\\
+METEOR& 71.20 & 89.63 & 2.07 & 87.62\\
+ROUGE & 70.76 & 90.04 & 2.00 & 87.71\\
+SPICE & \underline{71.94} & 88.80 & \underline{2.09} & 87.71\\
+BERTScore & 71.11 & 89.63 & 2.01 & 87.25\\
\hline
+GEST-SM& \textbf{72.89} & \textbf{90.87} & \textbf{2.21} & \textbf{89.80}\\
+GEST-NGM& 71.91 & \underline{90.46} & 2.05 & \underline{88.58}\\

\hline
\end{tabular}
\end{center}
\caption{Results comparing the power of BLEURT coupled with well-known text similarity metrics and GEST, applied on stories from Videos-to-Paragraphs test set. Text metrics are computed on the ground truth stories, while GESTs are generated with a transformer learned on the training set. Same notations as in Tab.~\ref{table:metrics_story_vs_classic}.}
\label{table:metrics_story_learned}
\end{table}

\section{GEST Video Generation Engine}
\label{sec:GEST_engine}

\begin{figure}[h]
    \centering
    \includegraphics[width=\columnwidth]{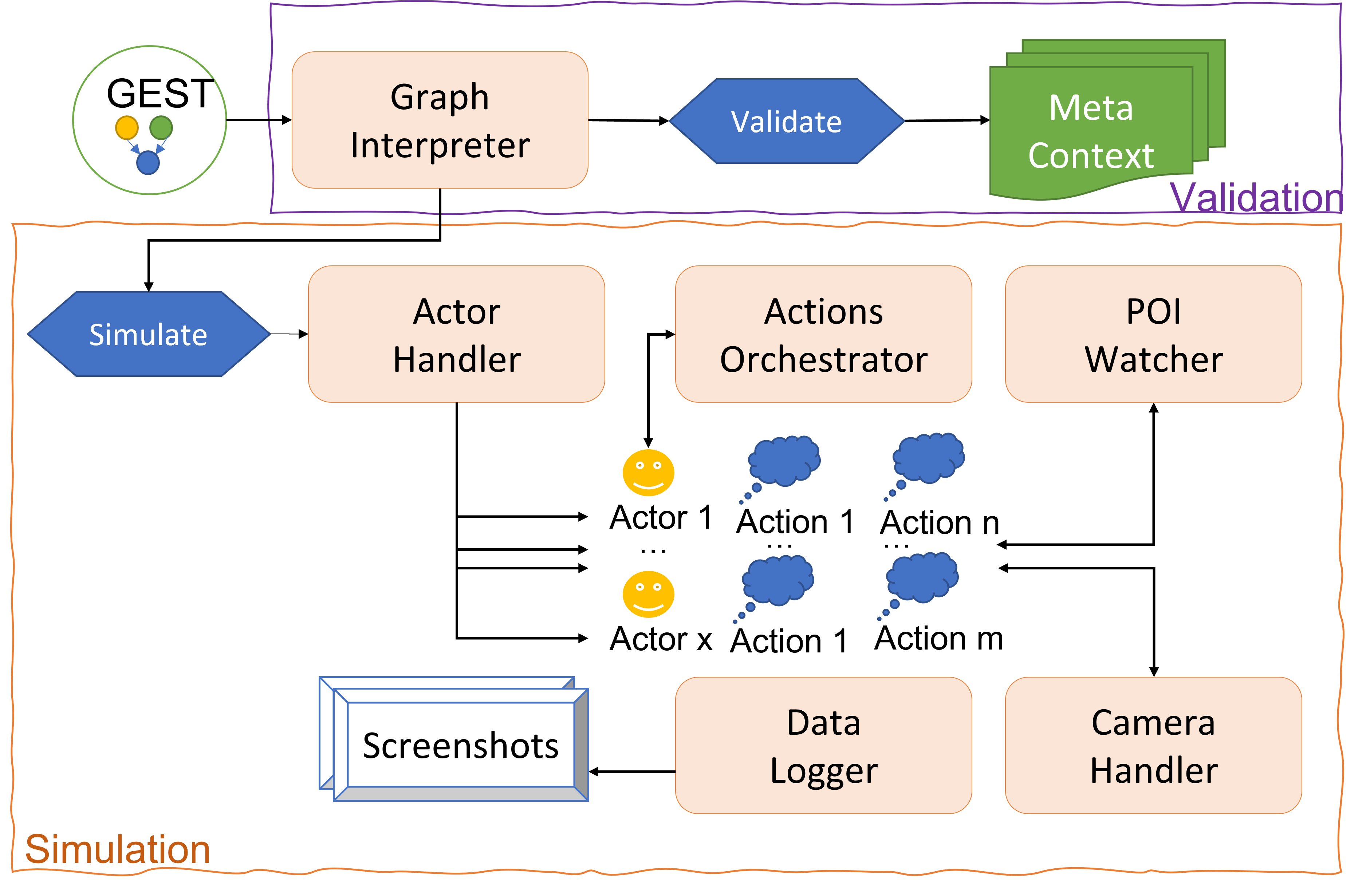}
    \caption{The system architecture of the engine. Upper part - meta context validation. Lower part - simulation. 
    }
    \label{fig:3dsysarch}
\end{figure}

To complete the connection between GEST and the visual world, we introduce the engine of visual stories. Based on the game GTA San Andreas with Multi Theft Auto (MTA)\footnote{\url{https://multitheftauto.com/}, accessed on 25 July 2023} interfacing the game's mechanics, we use the preexisting in-game locations, objects and animations and focus on events taking place in and around a house. 
The engine has full control within the virtual environment and can, therefore, take full advantage of the structured and explainable nature of GEST. It is capable of choosing a setting in a virtual environment, with locations, actions and entities that match the events described within the GEST and orchestrate the complex interactions during the simulation, thus emulating an entire world (Figure \ref{fig:3dsysarch}).

The system takes a GEST as input and, based on it, generates multiple valid videos - note the one-to-many relation. 
This engine is used to automatically generate videos from GEST. We couple this with the system that generates GEST starting from a text, closing the loop and building a system that transforms a text into a GEST, then a GEST into a video. We generate a set of 25 complex videos of 2-3 minutes each, with up to 15 different activities, much larger than what is used in the current literature. Even if the set is small, it is very challenging so we use to validate the quality of the generated videos. Results of this evaluation are presented in the following section.

\begin{table}[h]
\begin{center}
\begin{tabular}{|l|c|c|c|}
\hline
Metric & Ours & CogVideo & Text2VideoZero\\
\hline\hline
Bleu@4\cite{papineni-etal-2002-bleu}& 9.84 & 8.16 & \textbf{10.02} \\
Meteor\cite{banerjee-lavie-2005-meteor}& \textbf{14.16} & 13.48 & 13.96 \\
ROUGE\cite{lin-2004-rouge}& \textbf{35.40} & 32.72 & 34.87 \\
SPICE\cite{anderson2016spice}& \textbf{20.04} & 19.54 & 19.43 \\
CIDEr\cite{vedantam2015cider}& \textbf{34.12} & 33.16 & 33.65 \\
BERTScore\cite{zhang2019bertscore}& \textbf{19.37} & 13.09 & 15.02 \\
BLEURT\cite{sellam-etal-2020-bleurt}& \textbf{39.44} & 37.55 & 38.40 \\


\hline
\end{tabular}
\end{center}
\caption{Results on video-to-text task. We show in \textbf{bold} the best value for each metric.}
\label{table:video-to-text}
\end{table}

\section{Vision-Language Experiments with GEST}
\label{sec:vision_language_tests}

\begin{figure}
    \centering
    \includegraphics[width=0.6\columnwidth]{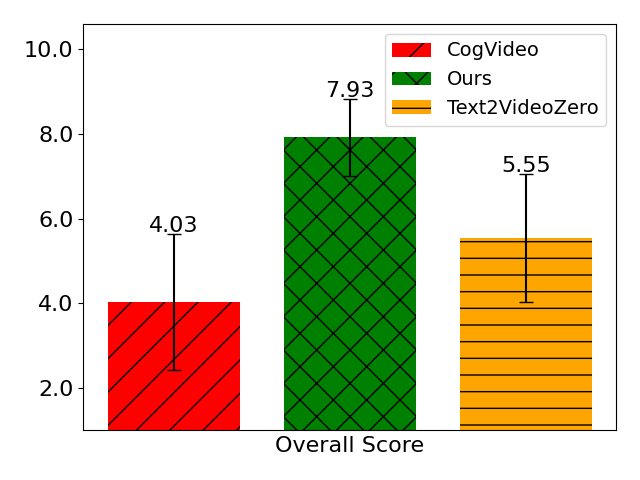}
    \caption{Overall scores (1-10) given by human evaluators.}
    \label{fig:human_overall}
\end{figure}

Next we present both human and automatic evaluations of our GEST-generated videos, compared to recent text-to-video models~\cite{hong2022cogvideo, khachatryan2023text2video}.  We invite human annotators to rate videos in terms of semantic content w.r.t input text, on a scale from 1 to 10 and pick the best video for each input text. We collected a total of 111 annotations, from 6 independent annotators.
In Fig~\ref{fig:human_overall} we show the overall scores given by human evaluators for each method.
In 87.39\% of cases our GEST-generated video was picked as best, with only 11.71\% for Text2VideoZero
and 0.90\% for CogVideo.

For the automatic evaluation of the generated videos, we use a state-of-the-art video-to-text generation method, VALOR~\cite{chen2023valor},  
and measure how well the text generated back from the generated videos match the initial input texts. VALOR is trained and tested separately for each type of video generation method using 5-fold cross validation, from scratch, over 3 runs with results averaged (shown in Tab. \ref{table:video-to-text}). 
These experiments match the human evaluation, keeping the same ranking across methods and proving that GEST-generated videos can better maintain the semantic content of the original input text. This proves that an explicit and fully explainable vision-language model in the form of a graph of events in space and time, could also provide in practice a better way to explain and control semantic content - thus bringing a complementary value in the context 
of realistic (but not necessarily truthful) AI generation models.

The reason why current deep learning models are not strong is that we generate long and complex videos. Their main weakness resides in their inability to integrate long and complex context, both in video and text generation.


\begin{figure}[h]
    \centering
    \includegraphics[width=\columnwidth]{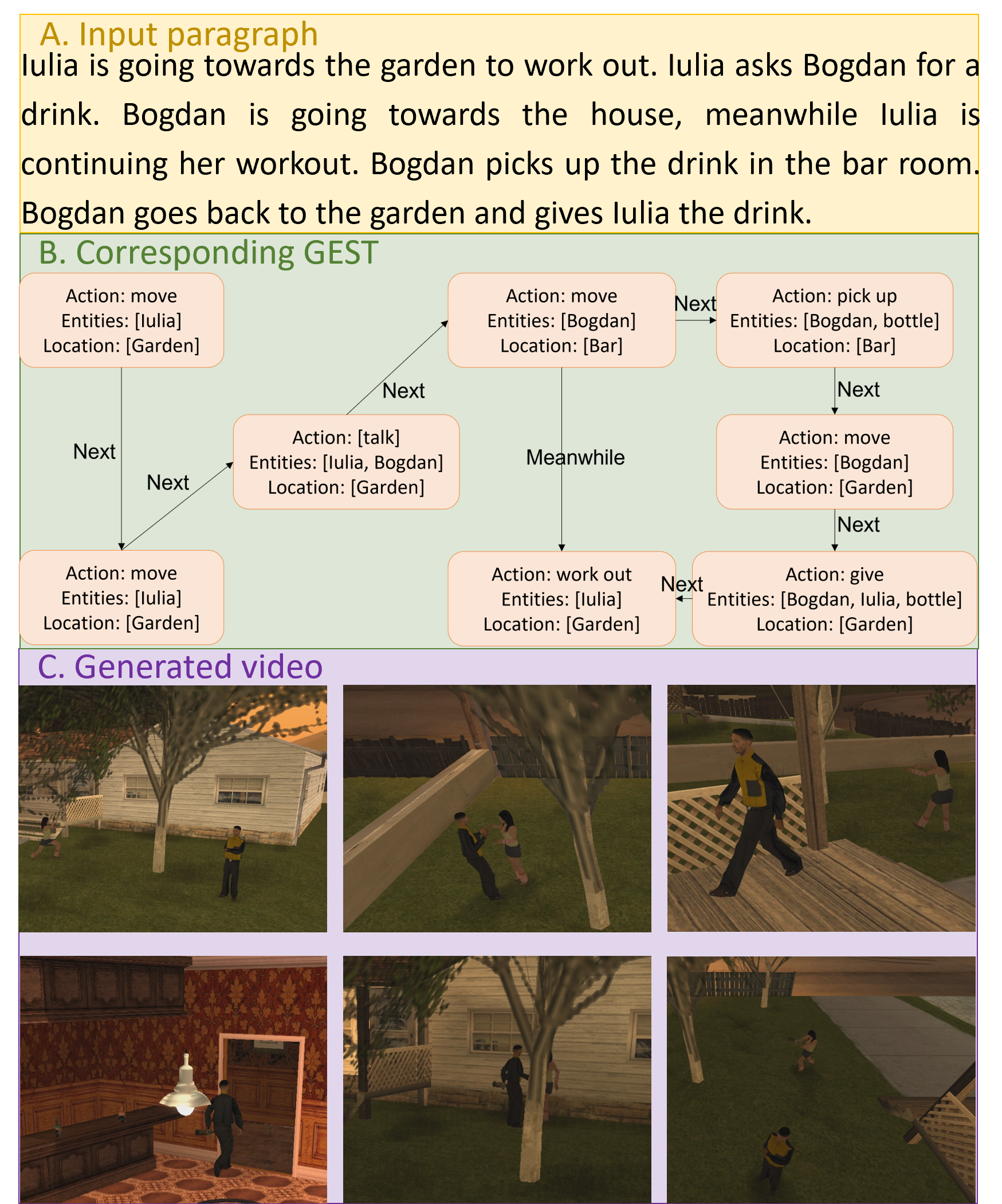}
    \caption{Example of input text (A), generated GEST from text (B) and automatically generated video from GEST (C).}
    \label{fig:samplevid}
\end{figure}

\section{Conclusions}

We propose an explainable representation that connects language and vision (see Fig~\ref{fig:general}), which explicitly captures semantic content as a graph of events in space and time (GEST). We prove that GEST is capable of capturing meaning from text and contribute to the design of powerful text-to-text comparison metrics when combined with graph matching. More importantly, GEST can be also used to generate videos from text that better preserve the semantic content (as evaluated by humans and automatic procedures), than deep learning methods for which there is no explicit way of explaining and controlling content. 
In future work we plan to explore ways to better integrate the power of deep learning into the explainable structure of GEST, for further developing a robust and trustworthy bridge between vision and language.

\textbf{Acknowledgements}: This work was funded in part by UEFISCDI, under Project EEA-RO-2018-0496 and by a Google Research Gift.

{\small
\bibliographystyle{ieee_fullname}
\bibliography{egbib}

\begin{thebibliography}{10}\itemsep=-1pt

\bibitem{anderson2016spice}
Peter Anderson, Basura Fernando, Mark Johnson, and Stephen Gould.
\newblock Spice: Semantic propositional image caption evaluation.
\newblock In {\em Computer Vision--ECCV 2016: 14th European Conference,
  Amsterdam, The Netherlands, October 11-14, 2016, Proceedings, Part V 14},
  pages 382--398. Springer, 2016.

\bibitem{antol2015vqa}
Stanislaw Antol, Aishwarya Agrawal, Jiasen Lu, Margaret Mitchell, Dhruv Batra,
  C Lawrence~Zitnick, and Devi Parikh.
\newblock Vqa: Visual question answering.
\newblock In {\em Proceedings of the IEEE international conference on computer
  vision}, pages 2425--2433, 2015.

\bibitem{balaji2019conditional}
Yogesh Balaji, Martin~Renqiang Min, Bing Bai, Rama Chellappa, and Hans~Peter
  Graf.
\newblock Conditional gan with discriminative filter generation for
  text-to-video synthesis.
\newblock In {\em IJCAI}, volume~1, page~2, 2019.

\bibitem{banarescu2013abstract}
Laura Banarescu, Claire Bonial, Shu Cai, Madalina Georgescu, Kira Griffitt, Ulf
  Hermjakob, Kevin Knight, Philipp Koehn, Martha Palmer, and Nathan Schneider.
\newblock Abstract meaning representation for sembanking.
\newblock In {\em Proceedings of the 7th linguistic annotation workshop and
  interoperability with discourse}, pages 178--186, 2013.

\bibitem{banerjee-lavie-2005-meteor}
Satanjeev Banerjee and Alon Lavie.
\newblock {METEOR}: An automatic metric for {MT} evaluation with improved
  correlation with human judgments.
\newblock In {\em Proceedings of the {ACL} Workshop on Intrinsic and Extrinsic
  Evaluation Measures for Machine Translation and/or Summarization}, pages
  65--72, Ann Arbor, Michigan, June 2005. Association for Computational
  Linguistics.

\bibitem{bogolin2020hierarchical}
Simion-Vlad Bogolin, Ioana Croitoru, and Marius Leordeanu.
\newblock A hierarchical approach to vision-based language generation: from
  simple sentences to complex natural language.
\newblock In {\em Proceedings of the 28th International Conference on
  Computational Linguistics}, pages 2436--2447, 2020.

\bibitem{brendel2011learning}
William Brendel and Sinisa Todorovic.
\newblock Learning spatiotemporal graphs of human activities.
\newblock In {\em 2011 International Conference on Computer Vision}, pages
  778--785. IEEE, 2011.

\bibitem{chen2023valor}
Sihan Chen, Xingjian He, Longteng Guo, Xinxin Zhu, Weining Wang, Jinhui Tang,
  and Jing Liu.
\newblock Valor: Vision-audio-language omni-perception pretraining model and
  dataset.
\newblock {\em arXiv preprint arXiv:2304.08345}, 2023.

\bibitem{cherian20222}
Anoop Cherian, Chiori Hori, Tim~K Marks, and Jonathan Le~Roux.
\newblock (2.5+ 1) d spatio-temporal scene graphs for video question answering.
\newblock In {\em Proceedings of the AAAI Conference on Artificial
  Intelligence}, volume~36, pages 444--453, 2022.

\bibitem{christensen2013towards}
Janara Christensen, Stephen Soderland, Oren Etzioni, et~al.
\newblock Towards coherent multi-document summarization.
\newblock In {\em Proceedings of the 2013 conference of the North American
  chapter of the association for computational linguistics: Human language
  technologies}, pages 1163--1173, 2013.

\bibitem{gao2017video}
Lianli Gao, Zhao Guo, Hanwang Zhang, Xing Xu, and Heng~Tao Shen.
\newblock Video captioning with attention-based lstm and semantic consistency.
\newblock {\em IEEE Transactions on Multimedia}, 19(9):2045--2055, 2017.

\bibitem{hong2022cogvideo}
Wenyi Hong, Ming Ding, Wendi Zheng, Xinghan Liu, and Jie Tang.
\newblock Cogvideo: Large-scale pretraining for text-to-video generation via
  transformers.
\newblock {\em arXiv preprint arXiv:2205.15868}, 2022.

\bibitem{khachatryan2023text2video}
Levon Khachatryan, Andranik Movsisyan, Vahram Tadevosyan, Roberto Henschel,
  Zhangyang Wang, Shant Navasardyan, and Humphrey Shi.
\newblock Text2video-zero: Text-to-image diffusion models are zero-shot video
  generators.
\newblock {\em arXiv preprint arXiv:2303.13439}, 2023.

\bibitem{leordeanu2005spectral}
Marius Leordeanu and Martial Hebert.
\newblock A spectral technique for correspondence problems using pairwise
  constraints.
\newblock In {\em Tenth IEEE International Conference on Computer Vision
  (ICCV'05) Volume 1}, volume~2, pages 1482--1489 Vol. 2, 2005.

\bibitem{li2018video}
Yitong Li, Martin Min, Dinghan Shen, David Carlson, and Lawrence Carin.
\newblock Video generation from text.
\newblock In {\em Proceedings of the AAAI conference on artificial
  intelligence}, volume~32, 2018.

\bibitem{lin-2004-rouge}
Chin-Yew Lin.
\newblock {ROUGE}: A package for automatic evaluation of summaries.
\newblock In {\em Text Summarization Branches Out}, pages 74--81, Barcelona,
  Spain, July 2004. Association for Computational Linguistics.

\bibitem{lin1998dependency}
Dekang Lin.
\newblock A dependency-based method for evaluating broad-coverage parsers.
\newblock {\em Natural Language Engineering}, 4(2):97--114, 1998.

\bibitem{lu2016hierarchical}
Jiasen Lu, Jianwei Yang, Dhruv Batra, and Devi Parikh.
\newblock Hierarchical question-image co-attention for visual question
  answering.
\newblock In {\em Advances in neural information processing systems}, pages
  289--297, 2016.

\bibitem{mann1988rhetorical}
William~C Mann and Sandra~A Thompson.
\newblock Rhetorical structure theory: Toward a functional theory of text
  organization.
\newblock {\em Text-interdisciplinary Journal for the Study of Discourse},
  8(3):243--281, 1988.

\bibitem{papineni-etal-2002-bleu}
Kishore Papineni, Salim Roukos, Todd Ward, and Wei-Jing Zhu.
\newblock {B}leu: a method for automatic evaluation of machine translation.
\newblock In {\em Proceedings of the 40th Annual Meeting of the Association for
  Computational Linguistics}, pages 311--318, Philadelphia, Pennsylvania, USA,
  July 2002. Association for Computational Linguistics.

\bibitem{pennington2014glove}
Jeffrey Pennington, Richard Socher, and Christopher~D. Manning.
\newblock Glove: Global vectors for word representation.
\newblock In {\em Empirical Methods in Natural Language Processing (EMNLP)},
  pages 1532--1543, 2014.

\bibitem{reed2016generative}
Scott Reed, Zeynep Akata, Xinchen Yan, Lajanugen Logeswaran, Bernt Schiele, and
  Honglak Lee.
\newblock Generative adversarial text to image synthesis.
\newblock In {\em International Conference on Machine Learning}, pages
  1060--1069. PMLR, 2016.

\bibitem{sellam-etal-2020-bleurt}
Thibault Sellam, Dipanjan Das, and Ankur Parikh.
\newblock {BLEURT}: Learning robust metrics for text generation.
\newblock In {\em Proceedings of the 58th Annual Meeting of the Association for
  Computational Linguistics}, pages 7881--7892, Online, July 2020. Association
  for Computational Linguistics.

\bibitem{singer2022make}
Uriel Singer, Adam Polyak, Thomas Hayes, Xi Yin, Jie An, Songyang Zhang, Qiyuan
  Hu, Harry Yang, Oron Ashual, Oran Gafni, et~al.
\newblock Make-a-video: Text-to-video generation without text-video data.
\newblock {\em arXiv preprint arXiv:2209.14792}, 2022.

\bibitem{singh2017graph}
Dinesh Singh and C~Krishna Mohan.
\newblock Graph formulation of video activities for abnormal activity
  recognition.
\newblock {\em Pattern Recognition}, 65:265--272, 2017.

\bibitem{sridhar2010relational}
Muralikrishna Sridhar, Anthony~G Cohn, and David~C Hogg.
\newblock Relational graph mining for learning events from video.
\newblock {\em STAIRS 2010}, pages 315--327, 2010.

\bibitem{vaswani2017attention}
Ashish Vaswani, Noam Shazeer, Niki Parmar, Jakob Uszkoreit, Llion Jones,
  Aidan~N Gomez, {\L}ukasz Kaiser, and Illia Polosukhin.
\newblock Attention is all you need.
\newblock In {\em Advances in neural information processing systems}, pages
  5998--6008, 2017.

\bibitem{vedantam2015cider}
Ramakrishna Vedantam, C Lawrence~Zitnick, and Devi Parikh.
\newblock Cider: Consensus-based image description evaluation.
\newblock In {\em Proceedings of the IEEE conference on computer vision and
  pattern recognition}, pages 4566--4575, 2015.

\bibitem{villegas2022phenaki}
Ruben Villegas, Mohammad Babaeizadeh, Pieter-Jan Kindermans, Hernan Moraldo,
  Han Zhang, Mohammad~Taghi Saffar, Santiago Castro, Julius Kunze, and Dumitru
  Erhan.
\newblock Phenaki: Variable length video generation from open domain textual
  description.
\newblock {\em arXiv preprint arXiv:2210.02399}, 2022.

\bibitem{wang2018reconstruction}
Bairui Wang, Lin Ma, Wei Zhang, and Wei Liu.
\newblock Reconstruction network for video captioning.
\newblock In {\em Proceedings of the IEEE conference on computer vision and
  pattern recognition}, pages 7622--7631, 2018.

\bibitem{wang2021neural}
Runzhong Wang, Junchi Yan, and Xiaokang Yang.
\newblock Neural graph matching network: Learning lawler’s quadratic
  assignment problem with extension to hypergraph and multiple-graph matching.
\newblock {\em IEEE Transactions on Pattern Analysis and Machine Intelligence},
  2021.

\bibitem{wang2018videos}
Xiaolong Wang and Abhinav Gupta.
\newblock Videos as space-time region graphs.
\newblock In {\em Proceedings of the European conference on computer vision
  (ECCV)}, pages 399--417, 2018.

\bibitem{wu2022nuwa}
Chenfei Wu, Jian Liang, Lei Ji, Fan Yang, Yuejian Fang, Daxin Jiang, and Nan
  Duan.
\newblock N{\"u}wa: Visual synthesis pre-training for neural visual world
  creation.
\newblock In {\em Computer Vision--ECCV 2022: 17th European Conference, Tel
  Aviv, Israel, October 23--27, 2022, Proceedings, Part XVI}, pages 720--736.
  Springer, 2022.

\bibitem{yuan2017temporal}
Yuan Yuan, Xiaodan Liang, Xiaolong Wang, Dit-Yan Yeung, and Abhinav Gupta.
\newblock Temporal dynamic graph lstm for action-driven video object detection.
\newblock In {\em Proceedings of the IEEE international conference on computer
  vision}, pages 1801--1810, 2017.

\bibitem{zettlemoyer2012learning}
Luke~S Zettlemoyer and Michael Collins.
\newblock Learning to map sentences to logical form: Structured classification
  with probabilistic categorial grammars.
\newblock {\em arXiv preprint arXiv:1207.1420}, 2012.

\bibitem{zhang2023language}
Muru Zhang, Ofir Press, William Merrill, Alisa Liu, and Noah~A Smith.
\newblock How language model hallucinations can snowball.
\newblock {\em arXiv preprint arXiv:2305.13534}, 2023.

\bibitem{zhang2019bertscore}
Tianyi Zhang, Varsha Kishore, Felix Wu, Kilian~Q Weinberger, and Yoav Artzi.
\newblock Bertscore: Evaluating text generation with bert.
\newblock {\em arXiv preprint arXiv:1904.09675}, 2019.

\bibitem{zhong2020self}
Huasong Zhong, Jingyuan Chen, Chen Shen, Hanwang Zhang, Jianqiang Huang, and
  Xian-Sheng Hua.
\newblock Self-adaptive neural module transformer for visual question
  answering.
\newblock {\em IEEE Transactions on Multimedia}, 23:1264--1273, 2020.

\bibitem{zhou2018end}
Luowei Zhou, Yingbo Zhou, Jason~J Corso, Richard Socher, and Caiming Xiong.
\newblock End-to-end dense video captioning with masked transformer.
\newblock In {\em Proceedings of the IEEE Conference on Computer Vision and
  Pattern Recognition}, pages 8739--8748, 2018.

\bibitem{zhou2019text}
Xingran Zhou, Siyu Huang, Bin Li, Yingming Li, Jiachen Li, and Zhongfei Zhang.
\newblock Text guided person image synthesis.
\newblock In {\em Proceedings of the IEEE/CVF Conference on Computer Vision and
  Pattern Recognition}, pages 3663--3672, 2019.

\end{thebibliography}
}

\end{document}